\documentclass{article} 
\usepackage{iclr2026_conference,times}

\usepackage{hyperref}
\usepackage{url}
\usepackage{amsmath}
\usepackage{algorithm}
\usepackage{algorithmic}
\usepackage{amssymb}

\usepackage{xcolor}
\usepackage{colortbl}
\usepackage{booktabs}
\usepackage{multirow}
\usepackage{graphicx}
\usepackage{amsmath}
\usepackage{amsthm}
\usepackage[switch]{lineno}
\usepackage{tabularray}
\usepackage[export]{adjustbox}
\usepackage{adjustbox}
\usepackage{array}

\usepackage{wrapfig}
\usepackage{booktabs}
\usepackage{graphicx}
\usepackage{amssymb}
\usepackage{tabularx}

\usepackage{cleveref}

\usepackage{pifont}  
\definecolor{my_green}{RGB}{51,102,0}
\definecolor{my_red}{RGB}{204, 0, 0}

\newcommand{\colorcmark}{\textcolor{my_green}{\ding{51}}}    

\usepackage[most]{tcolorbox}
\usepackage[table,xcdraw]{xcolor}

\definecolor{mybackcolor1}{RGB}{242,242,242}
\definecolor{myframecolor1}{RGB}{219,219,219}

\definecolor{windred}{RGB}{192,0,0}

\newtheorem{prompt}{Prompt}

\tcbset{
    prompt/.style={
        enhanced,
        breakable,
        colback=mybackcolor1,
        colframe=myframecolor1,
        boxrule=0mm,
        arc=0mm,
        before skip=10pt,
        after skip=10pt,
        frame hidden,
        title=#1,
        coltitle=black,
        fonttitle=\bfseries,
        colbacktitle=myframecolor1,
        width=\columnwidth,
        left=2mm,
        right=2mm,
        top=1mm,
        bottom=1mm,
    }
}

\title{Learn Before You Judge: Progressive Knowledge-to-Decision Alignment for Explainable Hateful Meme Detection}

\author{Bo Xu\textsuperscript{1}, 
Chenyuan Wang\textsuperscript{1}, 
Xinyu Chen\textsuperscript{1}, 
Quanhao Zhu\textsuperscript{1}, 
Rui Lin\textsuperscript{1}, \\
\textbf{Liang Zhao}\textsuperscript{1}, \textbf{Hongfei Lin}\textsuperscript{2}, \textbf{Feng Xia}\textsuperscript{3}\thanks{Corresponding authors} \\
\textsuperscript{1}School of Software, Dalian University of Technology,\\
\textsuperscript{2}School of Computer Science and Technology, Dalian University of Technology,\\
\textsuperscript{3}School of Computing Technologies, RMIT University,\\
\small{\texttt{boxu@dlut.edu.cn, chenyuanwang@mail.dlut.edu.cn, f.xia@ieee.org,}}\\
\small{\texttt{liangzhao@dlut.edu.cn}}\\
}

\iclrfinalcopy  
\begin{document}

\maketitle

\begin{abstract}
Hateful memes spread abusive content through implicit interactions between images and text, posing serious threats to the safety of online communities. In recent years, multimodal large language models have been widely used for hateful meme detection and are increasingly adopted to generate explainable detection results. However, we find that existing explain-then-detect methods often couple explanation generation and label prediction within the same training process. This coupling causes interference between task objectives, leading to limited detection performance and even worse results than simple SFT baselines. To address these challenges, we propose ProKDA, a progressive knowledge-to-decision alignment method for explainable hateful meme detection. Inspired by the human annotation training process, ProKDA first uses an agentic background knowledge construction pipeline to obtain external knowledge related to meme understanding. It then adopts a three-stage training strategy that sequentially performs background knowledge learning, hatefulness detection learning, and hatefulness boundary alignment. Unlike prior explain-then-detect methods that jointly optimize both tasks, ProKDA focuses on a single training objective at each stage. This design reduces interference between the two tasks and progressively transforms background knowledge into robust detection decisions. Experiments on three public hateful meme benchmarks show that ProKDA achieves state-of-the-art detection performance and provides accurate, explainable, and evidence-supported decisions for hateful meme moderation. Project page: \url{https://meizhiyuan88666.github.io/prokda}.
\end{abstract}
\textit{\textcolor{windred}{Content warning: This paper includes examples of hateful or offensive content for research and analysis purposes, which may be disturbing to some readers.}}

\section{Introduction}

With the rapid development of the Internet and social media, memes have become a prevalent medium on social media platforms, usually consisting of images and text \citep{1-Dancygier}. Their concise, humorous, and highly shareable form makes them an important way for users to express opinions and emotions \citep{2-Lestari,3-Castao}. However, memes can also be used to incite hatred, manipulate public opinion, and spread prejudice or harmful values \citep{4-Thomas,5-Martinez,6-Kennedy}. Hateful memes not only harm individuals and targeted groups, but also threaten social stability and the health of online communities \citep{7-Alafnan,8-Sharma}. Therefore, accurate and automatic hateful meme detection is crucial.

In recent years, the release of several benchmarks has advanced research on hateful meme detection, such as the Hateful Memes Challenge dataset (HMC) \citep{9-hmc} and the Multimedia Automatic Misogyny Identification dataset (MAMI) \citep{10-mami}. Early studies mainly focus on binary classification, i.e., determining whether a meme is hateful or benign. These methods usually use textual and visual features from memes and improve detection performance through attention mechanisms or other multimodal fusion strategies \citep{11-aomd,12-MultiOFF,13-Hate-CLIPper,14-Giovanni}. However, this direct detection paradigm provides limited explainability, which limits its use in real-world scenarios.

\begin{figure*}[tb] 
\centering 
\includegraphics[width=0.97\textwidth]{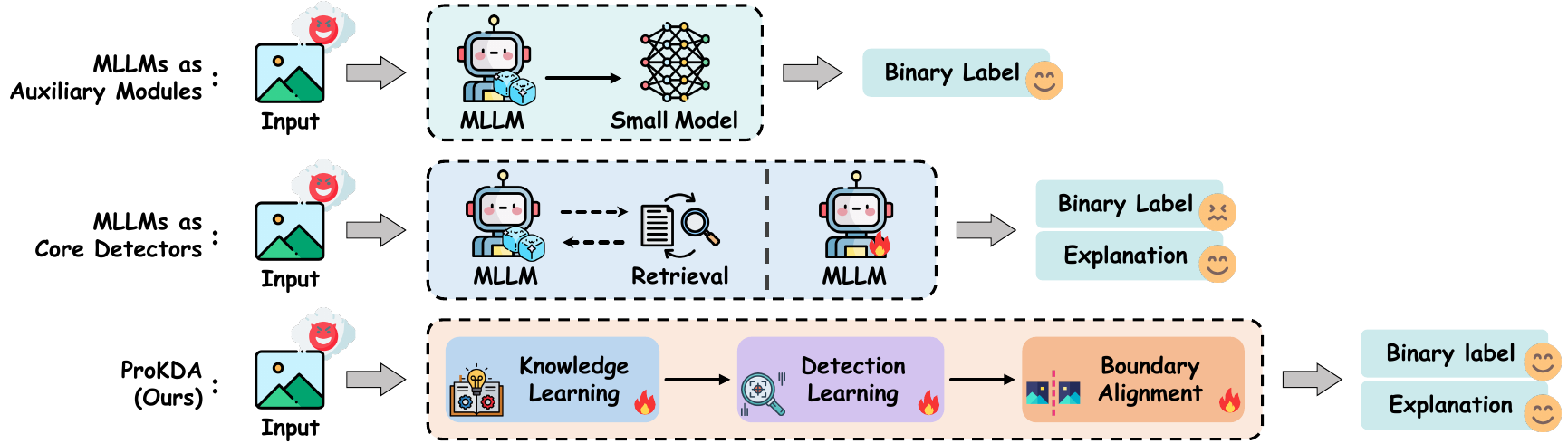} %
\caption{Comparison of ProKDA with existing MLLM-based methods.}
\label{fig:1-introduction}
\end{figure*}

The development of multimodal large language models (MLLMs) \citep{15-Liu,16-minigptv2,17-DeepSeek-vl,18-Qwen-VL,19-fuyu-8b,20-LLaVA-NeXT} provides new opportunities for hateful meme detection. Their visual understanding and cross-modal reasoning abilities enable them to generate explanations for hateful memes, which can assist human moderation and support human understanding of hateful content. However, memes often express hatred and abuse through metaphors \citep{21-metmeme,22-memecap}, sarcasm \citep{23-nykms}, and other implicit forms, making both detection and explainability challenging. Recent studies attempt to build an explain-then-detect reasoning paradigm through chain-of-thought reasoning \citep{24-expohm,25-Kmainasi,26-Pan}, in which models generate both explanations and detection labels. These methods further optimize classification accuracy and explanation quality through supervised fine-tuning (SFT) \citep{27-Ouyang} and reinforcement learning (RL) techniques such as Group Relative Policy Optimization (GRPO) \citep{28-deepseek-r1}. However, this paradigm essentially couples explanation generation and label prediction within the same training process. Such coupling causes interference between the two task objectives \citep{29-VC-Soup,30-Li}, thereby affecting both detection performance and explanation quality.

To address these issues, we propose ProKDA, a \textbf{\underline{Pro}}gressive \textbf{\underline{K}}nowledge-to-\textbf{\underline{D}}ecision \textbf{\underline{A}}lignment method. The method is inspired by the human annotation training process: annotators usually first study annotation guidelines and necessary background knowledge, then annotate samples, and further review ambiguous or difficult cases \citep{31-multimet}. Based on this idea, ProKDA first uses an agentic background knowledge construction pipeline to generate relevant background knowledge for each training meme. We then design a three-stage progressive training strategy. The first stage learns background knowledge and its relation to meme expressions, the second stage adapts the model to hateful meme classification through SFT, and the third stage applies confidence-based sample filtering and Direct Preference Optimization (DPO) \citep{33-dpo} to refine the detection boundary. Unlike existing explain-then-detect training paradigms, our strategy assigns a single training objective to each stage, reducing interference between explanation generation and label prediction. During DPO, we construct preference samples directly from binary labels without complex reward design, while confidence-based filtering reduces training samples and focuses the model on boundary cases. In summary, our contributions are as follows:

\begin{itemize}
    \item We propose a progressive knowledge-to-decision alignment strategy for explainable hateful meme detection, improving both detection performance and explanation quality.
    \item Our method simulates the training process of human annotators by progressively performing background knowledge learning, hatefulness detection learning, and DPO-based boundary alignment on confidence-filtered samples, mitigating task interference between explanation generation and label prediction.
    \item Experimental results show that our method outperforms both direct detection and explain-then-detect paradigms, achieving state-of-the-art performance on three hateful meme benchmarks.
\end{itemize}

\section{Related Work}

Existing MLLM-based methods can be grouped into two categories: auxiliary-module methods that provide explanations or knowledge for classifiers, and core-detector methods that directly perform reasoning and prediction. Their comparison with ProKDA is illustrated in Figure~\ref{fig:1-introduction}.

\paragraph{MLLMs as Auxiliary Modules.}

These methods typically use MLLMs as interpreters or knowledge generators to support downstream classifiers \citep{39-Mei,40-Mei,46-Tzelepi}. For example, \citet{34-Lin} distill multimodal reasoning knowledge from large models, while \citet{35-Towards} use MLLM-based debate to generate harmfulness rationales. \citet{36-Kumari} obtain multi-level meme analyses through multi-hop prompting, and \citet{37-Garg} convert MLLM-generated explanations into a knowledge graph for classifier enhancement. \citet{38-Ming} further use a frozen MLLM to generate explanatory analyses and combine them with meme representations for hateful meme detection. Although these studies improve detection performance, they do not provide direct explanations for meme predictions, leaving their explainability limited \citep{50-GOAT-Bench}.

\paragraph{MLLMs as Auxiliary Modules.}

Recent studies use MLLMs as core units for hateful meme reasoning through prompting, agent frameworks, or parameter-efficient fine-tuning \citep{47-Naquee,48-Olivia,49-M-QUEST}. For example, \citet{41-MemeGuard} combines an MLLM and an LLM for harmful meme understanding and intervention generation. \citet{42-Cao} use task-specific LoRA modules \citep{43-lora} for few-shot detection, while \citet{44-lorehm} and \citet{45-kdd} improve MLLM-based detection through retrieval, self-reflection, contrastive samples, or agent experience summarization. Recent studies further introduce RL for joint detection and explanation optimization. \citet{25-Kmainasi} use SFT followed by GRPO to optimize label prediction and explanation quality, while \citet{24-expohm} combine SFT warm-up, curriculum-based GRPO, and CDE rewards to further improve performance.

Although these methods use MLLMs as core detectors, they still have limitations. Prompting- and agent-based methods rely heavily on the native reasoning ability of MLLMs, making performance sensitive to model capacity and prompt design. SFT- and RL-based methods usually optimize explanation generation and label prediction together, which can cause task interference and make it difficult to balance detection performance and explanation quality \citep{51-xu}. To address these issues, we propose a progressive knowledge-to-decision alignment strategy that separates knowledge learning, detection learning, and boundary alignment into different stages, reducing task interference while improving both detection performance and explainability.

\section{Preliminaries}
\paragraph{Problem Statement.}
Given a hateful meme dataset $\mathcal{D}=\left\{\left( I_i, y_{i}^{*} \right) \right\} _{i=1}^{N}$
where $I_i$ denotes the $i$-th meme image and $y_i^*$ denotes the ground-truth label, we define hateful meme detection as a binary classification task. The model is required to generate a textual label $y_{i}^{*}\in \left\{\texttt{yes},\texttt{no} \right\}$, where $\texttt{yes}$ indicates hateful and $\texttt{no}$ indicates benign \citep{35-Towards}.

\paragraph{Supervised Fine-Tuning (SFT).}
Given an SFT training set $\mathcal{D}_{\mathrm{sft}}=\left\{\left( I_i, p_i,o_{i}^{*} \right) \right\} _{i=1}^{N}$, where $I_i$ denotes the $i$-th meme image, $p_i$ denotes the input prompt, and $o_{i}^{*}$ denotes the desired target text. Let the MLLM be $\pi _{\theta}$. SFT minimizes the negative log-likelihood of the target output:
\begin{equation}
    \mathcal{L} _{\mathrm{SFT}}(\theta )=-\mathbb{E} _{(I_i,p_i,o_{i}^{*})\sim \mathcal{D} _{\mathrm{sft}}}\sum_{t=1}^{|o_{i}^{*}|}{\log P_{\theta}(o_{i,t}^{*}}\mid o_{i,<t}^{*},I_i,p_i),
    \label{eq:eq1}
\end{equation}
where $o_{i,t}^{*}$ denotes the $t$-th token in the target output $o_{i}^{*}$, and $o_{i,<t}^{*}$ denotes the tokens before it.

\paragraph{Direct Preference Optimization (DPO).}
Given a preference dataset $\mathcal{D}_{\mathrm{dpo}}=\{(x_i,y_{i}^{w},y_{i}^{l})\}_{i=1}^{N}$, where $x_i$ denotes the input, $y_{i}^{w}$ denotes the preferred response, and $y_{i}^{l}$ denotes the non-preferred response. Let the reference model be $\pi _{\mathrm{ref}}$. DPO optimizes the model by comparing the relative probabilities of $y_{i}^{w}$ and $y_{i}^{l}$:
\begin{equation}
    \mathcal{L} _{\mathrm{DPO}}(\pi _{\theta};\mathcal{D} _{\mathrm{dpo}})=-\mathbb{E} _{(x_i,y_{i}^{w},y_{i}^{l})\sim \mathcal{D} _{\mathrm{dpo}}}\left[ \log \sigma \left( \beta \log \frac{\pi _{\theta}(y_{i}^{w}\mid x_i)}{\pi _{\mathrm{ref}}(y_{i}^{w}\mid x_i)}-\beta \log \frac{\pi _{\theta}(y_{i}^{l}\mid x_i)}{\pi _{\mathrm{ref}}(y_{i}^{l}\mid x_i)} \right) \right],
    \label{eq:eq2}
\end{equation}
where $\sigma(\cdot)$ is the sigmoid function and $\beta$ controls the preference strength.

\section{ProKDA Methodology}

\begin{figure*}[tb] 
\centering 
\includegraphics[width=0.98\textwidth]{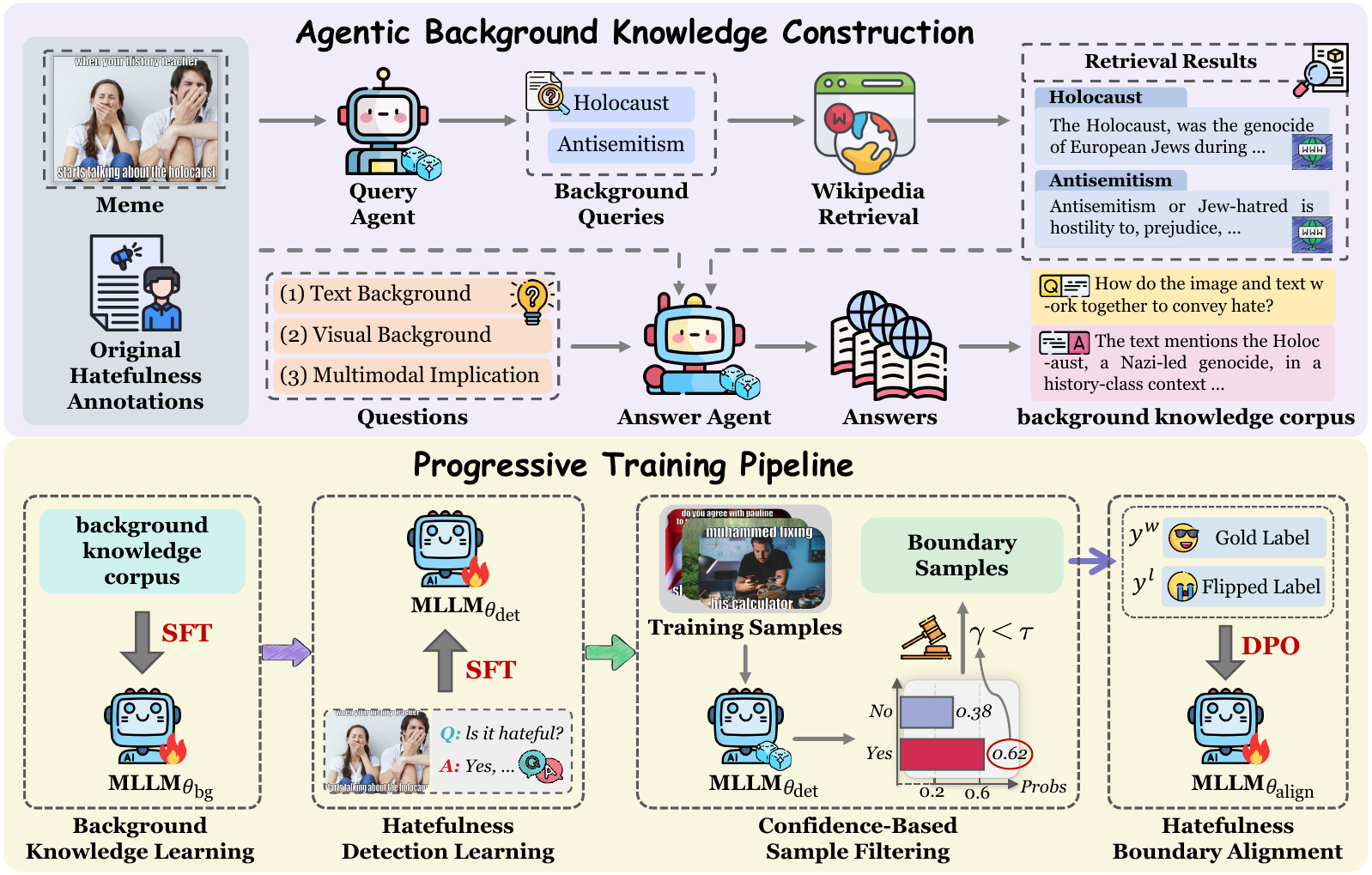} %
\caption{Overview of ProKDA. ProKDA first constructs a background knowledge corpus through an agentic pipeline, and then progressively optimizes the MLLM with knowledge learning, detection learning, confidence-based filtering, and DPO-based boundary alignment.}
\label{fig:2-method}
\end{figure*}

\subsection{Agentic Background Knowledge Construction}
\label{sec:4-1-agentic-background-knowledge-construct}

Memes often rely on external background knowledge, such as historical events, public figures, and cultural symbols. For example, understanding ``Holocaust'' in Figure~\ref{fig:2-method} requires knowledge of the event and its antisemitic context. Therefore, we design an agentic pipeline to construct background knowledge corpus for meme understanding.

Given the $i$-th meme sample, its image, text, and original hatefulness annotation are denoted as $I_i$, $T_i$, and $h_i$, respectively. Here, $h_i$ contains the hatefulness label and the attack target. First, a query agent $A_{\mathrm{q}}$ generates background knowledge queries for the current meme:
\begin{equation}
    S_i=A_{\mathrm{q}}(I_i, T_i, h_i).
    \label{eq:eq3}
\end{equation}
Here, $S_i$ contains one to two concise queries for retrieving key meme-related background concepts from Wikipedia:
\begin{equation}
    R_i=\mathrm{Retrieve}(S_i).
    \label{eq:eq4}
\end{equation}
We then use an answer agent $A_{\mathrm{a}}$ to convert retrieved content into structured question-answer pairs. $A_{\mathrm{a}}$ takes $I_i$, $T_i$, $h_i$, and $R_i$ as input, and answers three questions: what the text conveys, what the visual content conveys, and how the image and text jointly express the meaning. Let these three questions form a set $\mathcal{C} =\{c_{\mathrm{text}},c_{\mathrm{vis}},c_{\mathrm{mm}}\}$. The corresponding answer is:
\begin{equation}
    a_{i}^{j}=A_{\mathrm{a}}(I_i, T_i, h_i, R_i, c_j), c_j\in C.
    \label{eq:eq5}
\end{equation}
We compose the three answers into the background knowledge of the current meme:
\begin{equation}
    b_i=\mathrm{Compose}(a_{i}^{\mathrm{text}},a_{i}^{\mathrm{vis}}, a_{i}^{\mathrm{mm}}) .
    \label{eq:eq6}
\end{equation}
Here, $b_i$ denotes the background knowledge of the $i$-th meme, which connects textual and visual meanings to support the final judgment. The constructed background knowledge corpus is:
\begin{equation}
    \mathcal{D}_{\mathrm{bg}}=\left\{(I_i,T_i,h_i,b_i)\right\}_{i=1}^{N} .
    \label{eq:eq7}
\end{equation}

\subsection{Progressive Knowledge-to-Decision Training}

\paragraph{Background Knowledge Learning.}
Based on the background knowledge corpus $\mathcal{D}_{\mathrm{bg}}$, ProKDA first performs background knowledge learning on the initial MLLM $\theta_0$. For the $i$-th sample, we format its original annotation $h_i$ and background knowledge $b_i$ into the target response $y_i^{\mathrm{bg}}=\mathrm{Format}(h_i,b_i)$. The model takes the meme and the task prompt as input, and learns to generate $y_i^{\mathrm{bg}}$ through SFT. This stage enables the model to acquire meme-related background knowledge and understand how it supports the final hateful or non-hateful decision. The resulting model $\theta_{\mathrm{bg}}$ is used for subsequent hatefulness detection learning.

\paragraph{Hatefulness Detection Learning.}
Initialized from $\theta_{\mathrm{bg}}$, this stage uses SFT to adapt the model to the binary detection task, producing the hatefulness detection model $\theta_{\mathrm{det}}$. The model directly outputs a concise hatefulness judgment instead of detailed explanations, converting background understanding into detection capability for subsequent boundary-sample alignment.

\paragraph{Confidence-Based Sample Filtering.}
ProKDA selects boundary samples for subsequent DPO training based on model confidence. This process simulates the further decision review of ambiguous or uncertain samples in human annotation. We use the model $\theta_{\mathrm{det}}$ to perform inference on the whole training set, and identify low-confidence samples based on the prediction distribution over hateful and benign labels. Given a meme sample $M_i=(I_i,T_i)$, the model outputs the logits for hateful and benign judgments given the detection prompt $p_{\mathrm{det}}$:
\begin{equation}
    \mathbf{z}_i=F_{\theta_{\mathrm{det}}}(I_i, T_i, p_{\mathrm{det}}),
    \label{eq:eq8}
\end{equation}
where $\mathbf{z}_i=[z_i^{\mathrm{yes}},z_i^{\mathrm{no}}]$ denotes the logits for the two labels. The label probabilities are obtained by softmax:
\begin{equation}
    [p_{i}^{\mathrm{yes}}, p_{i}^{\mathrm{no}}]=Softmax([z_{i}^{\mathrm{yes}}, z_{i}^{\mathrm{no}}]).
    \label{eq:eq9}
\end{equation}
The prediction confidence of the model on sample $M_i$ is defined as:
\begin{equation}
    \gamma_i=max(p_{i}^{\mathrm{yes}},p_{i}^{\mathrm{no}}).
    \label{eq:eq10}
\end{equation}
We set $\tau\in[0.5,1]$. Samples with $\gamma_i\geqslant\tau$ are classified with high confidence, while samples with $\gamma_i<\tau$ are treated as boundary samples, since they lie near the current decision boundary. Therefore, we construct the boundary sample set as:
\begin{equation}
    \mathcal{D}_{\mathrm{bd}}=\{M_i\in \mathcal{D}_{\mathrm{train}}\mid \gamma_i<\tau \}.
    \label{eq:eq11}
\end{equation} 
$\mathcal{D}_{\mathrm{bd}}$ is used in the hatefulness boundary alignment stage, allowing the model to focus only on ambiguous samples that are difficult to judge reliably.

\paragraph{Hatefulness Boundary Alignment.}
Based on the boundary sample set $\mathcal{D}_{\mathrm{bd}}$, ProKDA applies DPO to refine the hatefulness decision boundary. This stage calibrates the model on low-confidence samples, improving its ability to handle ambiguous cases. Since hateful meme detection is a binary classification task, we directly construct preference pairs from gold labels. For each $M_i\in \mathcal{D}_{\mathrm{bd}}$, the gold-label response is used as the preferred response $y_i^w$, while the flipped-label response is used as the non-preferred response $y_i^l$. We then obtain the DPO training set:
\begin{equation}
    \mathcal{D}_{\mathrm{dpo}}=\left\{ \left( M_i,p_{\det},y_{i}^{w},y_{i}^{l} \right) \mid M_i\in \mathcal{D}_{\mathrm{bd}} \right\}.
    \label{eq:eq12}
\end{equation} 
Starting from $\theta_{\mathrm{det}}$, we train the model on $\mathcal{D}_{\mathrm{dpo}}$ with the DPO objective to obtain the final model $\theta_{\mathrm{align}}$. This stage further calibrates the decision boundary on low-confidence samples.

\section{Experiments}

\subsection{Experimental Setup}
\paragraph{Datasets and Tasks.}
We evaluate ProKDA on three meme classification datasets: HMC, MAMI, and PrideMM \citep{53-MemeCLIP}. HMC covers different types of hateful memes, MAMI focuses on misogynous memes, and PrideMM focuses on LGBTQ+-related stances. We conduct binary classification on all datasets, classifying each meme as hateful or benign. We also evaluate model-generated explanations on HMC, since it provides human explanation annotations through HatReD \citep{52-hatred}. Detailed dataset descriptions and statistics are provided in Appendix~\ref{appendix:data_stats}.

\paragraph{Evaluation Metrics.}
For classification, we follow existing work \citep{54-M2KE} and use accuracy, macro-F1, and weighted-F1 as metrics. For explanation quality, we use LLM-as-a-judge \citep{55-Wang,56-HARE} and human evaluation to assess model-generated explanations against human references. Detailed evaluation settings are provided in Appendix~\ref{appendix:prompt_eval}, and the results of explanation quality evaluation are reported in Section~\ref{sec:explain-eval}.

\subsection{Baseline Models}

Table~\ref{tab:main-result} compares ProKDA with a comprehensive set of baselines based on Qwen2.5-VL-3B and Qwen2.5-VL-7B \citep{57-Qwen2.5-VL}. We briefly describe the baseline settings below, with implementation details and ProKDA prompts provided in Appendices~\ref{appendix:exp_setup} and~\ref{appendix:prompt_prokda}.

\paragraph{SFT \& DPO.}
We build two SFT baselines to compare the direct detection and explain-then-detect paradigms. SFT (Explain-then-Detect) uses the background knowledge explanations constructed in Section~\ref{sec:4-1-agentic-background-knowledge-construct} and gold labels as target responses, while SFT (Direct Detection) uses only gold labels for direct binary prediction. For DPO, we use the gold-label and flipped-label responses as the preferred and non-preferred responses, respectively. All SFT and DPO baselines use the same hyperparameter settings as ProKDA.

\paragraph{MLLMs as auxiliary modules.}
We compare ProKDA with representative methods that use MLLMs to provide auxiliary reasoning or knowledge for detection. ExplainHM \citep{35-Towards} uses MLLM debates to generate reasoning for a small language model detector, while M2KE \citep{54-M2KE} uses multiple agents to obtain high-quality rationales as knowledge-enhanced information for the subsequent detector.

\paragraph{MLLMs as core detectors.}
We compare ProKDA with training-free and training-based methods that use MLLMs as core detectors. Training-free methods, including U-CoT+ \citep{26-Pan}, LoReHM \citep{44-lorehm}, MIND \citep{58-mind}, and ALARM \citep{45-kdd}, rely on prompting, retrieval, reflection, or agent experience summarization. Training-based methods, including MemeReason \citep{25-Kmainasi}, RA-HMD \citep{40-Mei}, and ExPO-HM \citep{24-expohm}, further optimize MLLMs through SFT, RL, or reasoning-quality rewards. We exclude closed-source models such as the OpenAI series because safety filtering may block responses to highly harmful memes.

\definecolor{myrowcolor}{RGB}{242,235,249} 
\begin{table*}[!tb]

    \centering
    \small
    \caption{Comparison between ProKDA and baseline methods. Accuracy (ACC), macro-F1 (M-F1), and weighted-F1 (W-F1) are used as evaluation metrics (\%). Bold and underlined values denote the best and second-best results, respectively.}
    \begin{center}

    \setlength\tabcolsep{4pt}
    \resizebox{0.99\linewidth}{!}{
        \begin{tabular}{cl|ccc|ccc|ccc} 
        \toprule
        \multirow{2}{*}{\#} & \multirow{2}{*}{Model}              & \multicolumn{3}{c|}{HMC}                         & \multicolumn{3}{c|}{MAMI}                        & \multicolumn{3}{c}{PrideMM}                       \\
                            &                                     & Acc            & M-F1           & W-F1           & Acc            & M-F1           & W-F1           & Acc            & M-F1           & W-F1            \\ 
        \midrule
                            & \multicolumn{10}{l}{\textit{MLLMs as Auxiliary Modules}}                                                                                                                                      \\ 
        \midrule
        1                   & ExplainHM (LLaVA-v1.5-13B)          & 75.60          & 75.39          & -              & -              & -              & -              & -              & -              & -               \\
        2                   & M2KE (LLaVA-v1.5-13B)               & 75.76          & 75.62          & -              & 75.85          & 75.71          & -              & -              & -              & -               \\ 
        \midrule
                            & \multicolumn{10}{l}{\textit{MLLMs as Core Detectors (Training-Free)}}                                                                                                                         \\ 
        \midrule
        3                   & U-CoT+ (Qwen2.5-14B)                & 72.50          & 72.41          & -              & 79.90          & 79.89          & -              & 71.60          & 71.37          & -               \\
        4                   & LoReHM (LLaVA-34B)                  & 65.60          & 65.59          & -              & 75.40          & 75.28          & -              & -              & -              & -               \\
        5                   & MIND (LLaVA-v1.6-34B)               & 66.40          & 68.38          & -              & 73.60          & 75.38          & -              & -              & -              & -               \\
        6                   & ALARM (Qwen2.5-VL-72B)              & 75.80          & 75.79          & -              & \underline{81.28}  & 81.25          & -              & -              & -              & -               \\ 
        \midrule
                            & \multicolumn{10}{l}{\textit{MLLMs as Core Detectors (Training-Based)}}                                                                                                                        \\ 
        \midrule
        7                   & MemeReason~(Qwen3-VL-8B)            & \underline{81.20}  & 79.00          & \underline{81.00}  & -              & -              & -              & -              & -              & -               \\
        8                   & RA-HMD (Qwen2.5-VL-7B)              & 80.80          & 80.10          & -              & 81.00          & 81.00          & -              & \underline{78.00}  & 77.80          & -               \\
        9                   & ExPO-HM (Qwen2.5-VL-7B)             & -              & \underline{81.10}  & -              & -              & \underline{82.30}  & -              & -              & \underline{78.70}  & -               \\
        10                  & Qwen2.5-VL-3B                       &                &                &                &                &                &                &                &                &                 \\
        11                  & ~\textit{~Zero-shot}                 & 63.10          & 61.99          & 62.12          & 61.80          & 57.44          & 57.44          & 65.09          & 64.24          & 64.38           \\
        12                  & ~\textit{~SFT (Explain-then-Detect)} & 67.00          & 65.22          & 65.38          & 69.70          & 68.63          & 68.63          & 66.86          & 66.81          & 66.84           \\
        13                  & ~\textit{~SFT (Direct Detection)}    & 72.20          & 71.07          & 71.18          & 77.50          & 76.99          & 76.99          & 74.36          & 74.35          & 74.36           \\
        14                  & ~\textit{~DPO}                       & 71.60          & 70.24          & 70.36          & 76.40          & 75.65          & 75.65          & 74.75          & 74.75          & 74.73           \\
        15                  & ~\textit{~ProKDA}                    & 80.20          & 80.19          & 80.20          & 78.70          & 78.13          & 78.13          & 75.74          & 75.72          & 75.70           \\
        16                  & Qwen2.5-VL-7B                       &                &                &                &                &                &                &                &                &                 \\
        17                  & ~\textit{~Zero-shot}                 & 63.40          & 60.19          & 60.42          & 56.30          & 47.31          & 47.31          & 58.19          & 51.10          & 51.58           \\
        18                  & ~\textit{~SFT (Explain-then-Detect)} & 72.00          & 70.93          & 71.04          & 75.00          & 74.34          & 74.34          & 76.53          & 76.48          & 76.51           \\
        19                  & ~\textit{~SFT (Direct Detection)}    & 75.60          & 75.15          & 75.22          & 78.90          & 78.37          & \underline{78.37}  & 77.71          & 77.71          & \underline{77.72}   \\
        20                  & ~\textit{~DPO}                       & 74.30          & 73.65          & 73.73          & 77.30          & 76.60          & 76.60          & 76.53          & 76.53          & 76.53           \\
        \rowcolor{myrowcolor} 21                  & ~\textit{~ProKDA}                    & \textbf{83.50} & \textbf{83.41} & \textbf{83.44} & \textbf{83.10} & \textbf{83.08} & \textbf{83.08} & \textbf{80.28} & \textbf{80.27} & \textbf{80.28}  \\
        \bottomrule
        \end{tabular}
    }
    \label{tab:main-result}
    \end{center}
\end{table*}

\subsection{Comparing ProKDA to Baselines}
Table~\ref{tab:main-result} compares ProKDA with existing methods on HMC, MAMI, and PrideMM. Background knowledge quality analysis is provided in Appendix~\ref{appendix:background_quality}, and qualitative analysis and error cases are provided in Appendix~\ref{appendix:qualitative_ana}. We summarize the key observations below.

\paragraph{Explain-then-detect methods hurt classification performance.}

We first compare the two SFT baselines. SFT (Explain-then-Detect) consistently underperforms SFT (Direct Detection), indicating that jointly optimizing explanation generation and label prediction introduces task interference and hurts classification performance. Similarly, training-free explain-then-detect methods (\#3-\#6) also underperform the task-specific trained method RA-HMD. These results suggest that simply adding explicit rationales through CoT prompting or explanation-supervised post-training does not reliably improve hateful meme detection.

\paragraph{Task-specific training is more important than model scaling.}
Training-free methods generally underperform training-based methods when MLLMs are used as core detectors. Even ALARM with Qwen2.5-VL-72B underperforms ExPO-HM with Qwen2.5-VL-7B by 5.31 M-F1 on HMC, while zero-shot Qwen2.5-VL-3B/7B (\#11, \#17) also trail their SFT/DPO variants. This suggests that scaling alone cannot replace task-specific training.

\paragraph{Gold-label-only post-training shows a clear upper bound.}
SFT (Direct Detection) and DPO improve over zero-shot inference on all three datasets, but still lag behind stronger training-based methods such as MemeReason and RA-HMD. The small gap between SFT (Direct Detection) and DPO suggests that gold-label-only optimization reaches a limited performance ceiling. Moreover, these methods focus on final labels rather than faithful explanations for memes, limiting their explainability in real-world moderation.

\paragraph{ProKDA achieves stable improvements across datasets and model scales.}
ProKDA based on Qwen2.5-VL-7B achieves the best overall performance on all three datasets, outperforming ExPO-HM and other baselines. On HMC, it improves W-F1 by 10.9\% and 13.2\% over the corresponding SFT (Direct Detection) and DPO versions, respectively. ProKDA based on Qwen2.5-VL-3B also consistently outperforms its SFT and DPO counterparts and even surpasses ALARM on HMC. Increasing the model scale further strengthens ProKDA (\#15→\#21), confirming its effectiveness across datasets and model scales.

\subsection{Ablation Study of ProKDA Components}

We conduct an ablation study on the three training stages of ProKDA on HMC to analyze the contribution of each stage. The results are shown in Table~\ref{tab:abalation-study}.

\definecolor{myrowcolor}{RGB}{242,235,249} 

\begin{wraptable}{r}{0.7\textwidth}
    \centering
    \small
    \caption{Ablation study of ProKDA components.}
    \begin{center}
    \setlength\tabcolsep{4pt}
    \resizebox{0.98\linewidth}{!}{
        \begin{tabular}{rccc|ccc|ccc} 
        \toprule
           & \multicolumn{3}{c|}{Components} & \multicolumn{3}{c|}{Qwen2.5-VL-3B} & \multicolumn{3}{c}{Qwen2.5-VL-7B}  \\
        \# & Stage 1 & Stage 2 & Stage 3     & Acc   & M-F1  & W-F1               & Acc   & M-F1  & W-F1               \\ 
        \midrule
        1  & -       & -       & -           & 63.10 & 61.99 & 62.12              & 63.40 & 60.19 & 60.42              \\
        2  & \colorcmark       & -       & -           & 68.50 & 68.42 & 68.39              & 74.70 & 74.63 & 74.61              \\
        3  & -       & \colorcmark       & -           & 72.20 & 71.07 & 71.18              & 75.60 & 75.15 & 75.22              \\
        4  & \colorcmark       & \colorcmark       & -           & 76.50 & 76.02 & 76.09              & 81.90 & 81.76 & 81.79              \\
        5  & -       & \colorcmark       & \colorcmark           & 76.20 & 75.80 & 75.87              & 80.40 & 80.27 & 80.31              \\
        \rowcolor{myrowcolor} 6  & \colorcmark       & \colorcmark       & \colorcmark           & 80.20 & 80.19 & 80.20              & 83.50 & 83.41 & 83.44              \\
        \bottomrule
        \end{tabular}
    }
    \label{tab:abalation-study}
    \end{center}
\end{wraptable}

\paragraph{Background knowledge learning provides the basis for progressive detection.}
Background knowledge learning (Stage 1) provides a key foundation for subsequent progressive training. With only Stage 1, the model already outperforms zero-shot inference. Applying Stage 2 on top of Stage 1 (\#4) also performs better than using Stage 2 alone (\#3), with W-F1 gains of 4.91 and 6.57 for the 3B and 7B models, respectively. Similarly, introducing Stage 1 into the Stage 2+3 setting (\#6 vs. \#5) further improves W-F1 by 4.33 and 3.13. These results show that Stage 1 helps the model acquire meme-related background knowledge and connect it with implicit semantics, background references, and attack intentions, providing a stronger basis for hatefulness detection. This observation is also consistent with \citet{24-expohm}.

\paragraph{Hatefulness detection learning and boundary alignment refine the decision boundary.}
Hatefulness detection learning (Stage 2) adapts the model to the detection task. Using only Stage 2 improves W-F1 by 9.06 and 14.8 over zero-shot inference for the 3B and 7B models, while applying it after Stage 1 (\#4) brings gains of 7.70 and 7.18. Boundary alignment (Stage 3) further improves performance, with gains of 4.11 and 1.65 when added on top of Stage 1+2 (\#6 vs. \#4). These results show that Stage 3 refines the decision boundary and helps the model handle ambiguous samples.

\subsection{Analysis of Confidence-based Sample Filtering}

We analyze confidence-based sample filtering through threshold effects and training-set confidence distributions. Additional analyses on confidence-interval accuracy and qualitative examples are provided in Appendix~\ref{appendix:confidence-filter}.

\begin{wrapfigure}{r}{0.65\textwidth}
\centering 
\includegraphics[width=0.64\textwidth]{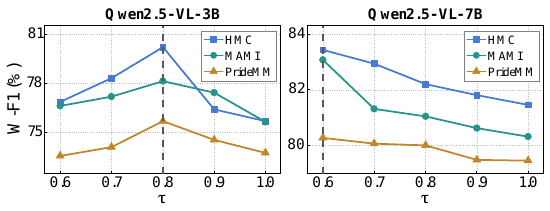} %
\caption{Performance under different confidence thresholds.}
\label{fig:4-effect-thresholds}
\end{wrapfigure}

\paragraph{Effect of Confidence Thresholds on Boundary Alignment.}

Figure~\ref{fig:4-effect-thresholds} shows the results under different confidence thresholds $\tau$. The optimal threshold depends on model scale: Qwen2.5-VL-3B performs best at $\tau=0.8$, while Qwen2.5-VL-7B performs best at $\tau=0.6$. We believe this is related to how well the model learns the training samples during hatefulness detection learning. For the 7B model, most samples already have confidence scores much higher than 0.8, indicating that it has sufficiently learned most samples and Stage 3 only needs to focus on low-confidence boundary samples. In contrast, Qwen2.5-VL-3B has a more dispersed confidence distribution, suggesting that the smaller model still benefits from including more medium-confidence samples for boundary alignment. Moreover, using all samples for Stage 3 often degrades performance by distracting the model from difficult cases. Confidence-based filtering instead preserves ambiguous boundary samples for more effective alignment.

\begin{figure*}[htb] 
\centering 
\includegraphics[width=0.98\textwidth]{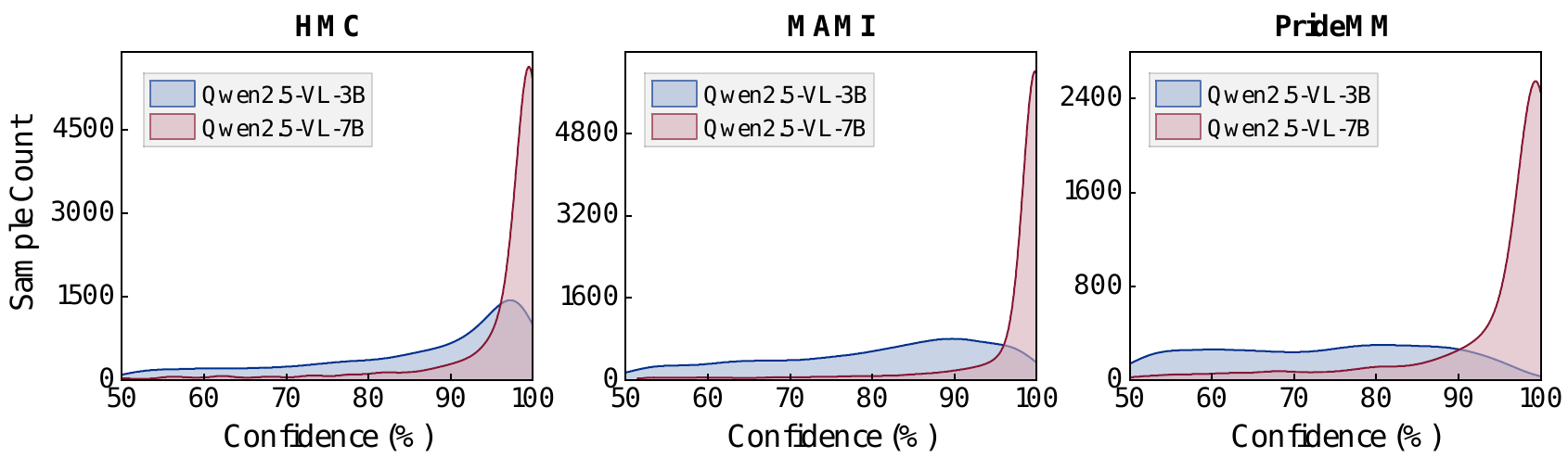} %
\caption{Model confidence distributions across different training sets.}
\label{fig:5-confidence-distribute}
\end{figure*}

\paragraph{Model Confidence Distributions on Training Sets.}
Figure~\ref{fig:5-confidence-distribute} shows the confidence distributions of Qwen2.5-VL-3B and Qwen2.5-VL-7B on the three training sets. The 7B model mainly concentrates in the [90,100] interval, while the 3B model is more dispersed. This suggests that the 7B model learns most samples more thoroughly, making low-confidence samples more indicative of boundary uncertainty. In contrast, the dispersed distribution of the 3B model suggests the need to include more medium-confidence samples for boundary alignment.

\subsection{Explainability Evaluation}
\label{sec:explain-eval}

\definecolor{myrowcolor}{RGB}{239,229,247} 

\begin{table*}[!htb]
    \small \centering
    \caption{Explanation evaluation of hateful memes, including Informativeness (Info.), Readability (Read.), Soundness (Sound.), and Persuasiveness (Pers.).}
    \begin{center}
    \setlength\tabcolsep{4pt}
    \resizebox{0.90\linewidth}{!}{
        \begin{tabular}{l|ccc|ccc|ccc} 
        \toprule
               & \multicolumn{3}{c|}{GPT-4o mini} & \multicolumn{3}{c|}{GPT-5} & \multicolumn{3}{c}{Human}   \\
               & Qwen & w/ Stage 1 & ProKDA       & Qwen & w/ Stage 1 & ProKDA & Qwen & w/ Stage 1 & ProKDA  \\ 
        \midrule
        Info.  & 2.93 & 4.08       & 4.25         & 2.20 & 3.62       & 3.83   & 2.35 & 3.74       & 3.94    \\
        Read.  & 4.34 & 4.62       & 4.78         & 4.50 & 4.73       & 4.87   & 4.47 & 4.75       & 4.79    \\
        Sound. & 2.83 & 4.05       & 4.23         & 1.98 & 3.33       & 3.56   & 2.14 & 3.51       & 3.67    \\
        Pers.  & 2.73 & 4.05       & 4.21         & 2.08 & 3.53       & 3.74   & 2.39 & 3.82       & 4.06    \\
        \bottomrule
        \end{tabular}
    }
    \label{tab:explain-result}
    \end{center}
\end{table*}

In this section, we evaluate the quality of model-generated meme explanations. Since textual explanations are diverse and have no fixed standard answer, we follow \citep{35-Towards,54-M2KE} and use both LLM-as-a-judge and human evaluation. We evaluate four dimensions: informativeness, readability, soundness, and persuasiveness, each scored on a 5-point Likert scale (see Appendix~\ref{appendix:explan_eval} for detailed prompts). GPT-4o mini and GPT-5 serve as LLM judges, while three computer science master's students with experience in hateful meme annotation conduct human evaluation. We evaluate Qwen2.5-VL-7B on the HMC test set with human gold explanation annotations \citep{52-hatred}, comparing the original model, the model after background knowledge learning (w/ Stage 1), and ProKDA. The results are shown in Table~\ref{tab:explain-result}.

\paragraph{ProKDA consistently improves the quality of model-generated explanations.}
Based on Table~\ref{tab:explain-result}, we make the following observations. First, w/ Stage 1 outperforms the original model on all four dimensions, showing that background knowledge learning improves meme understanding. Second, both w/ Stage 1 and ProKDA achieve clear gains in informativeness, soundness, and persuasiveness, indicating that the agentic background knowledge construction pipeline provides useful contextual evidence for explaining implicit hateful meanings. Third, the improvement in readability is relatively small, because the original model already has strong text generation ability. Finally, ProKDA further outperforms w/ Stage 1 on all four dimensions. This suggests that background knowledge alone is not enough: without explicit hatefulness detection learning, explanations may fail to focus on hateful intent. ProKDA addresses this by aligning background knowledge with detection decisions, making explanations more consistent with hateful intent and human references.

\section{Conclusion}

We propose ProKDA, combining agentic knowledge construction and progressive fine-tuning for explainable hateful meme detection. Experiments show that ProKDA achieves strong detection and explanation performance across datasets. Limitations and future work are in Appendix~\ref{appendix:future_work}.

\section*{Reproducibility Statement}

We provide the LLM usage statement in Appendix~\ref{appendix:llm_state}. Detailed dataset usage and statistics are provided in Appendix~\ref{appendix:data_stats}. The experimental settings, implementation details, software versions, hardware configurations, and hyperparameter settings are described in Appendix~\ref{appendix:exp_setup} to support reproducibility. The prompts used in ProKDA and the evaluation prompts are provided in Appendices~\ref{appendix:prompt_prokda} and~\ref{appendix:prompt_eval}, respectively. Additional analyses, including background knowledge quality analysis, confidence-based sample filtering analysis, and qualitative analysis, are reported in Appendices~\ref{appendix:background_quality},~\ref{appendix:confidence-filter}, and~\ref{appendix:qualitative_ana}. Limitations, future work, and ethical considerations are discussed in Appendices~\ref{appendix:future_work} and~\ref{appendix:ethical_state}.

\bibliography{iclr2026_conference}
\bibliographystyle{iclr2026_conference}

\appendix
\section{LLM Usage Statement}
\label{appendix:llm_state}
In the process of developing this work, we utilized LLMs for grammar correction and occasionally as a rewriting tool.

\section{Dataset Details and Statistics}
\label{appendix:data_stats}
\subsection{Dataset Details}

\paragraph{HMC.}
The Hateful Memes Challenge (HMC) \citep{9-hmc} is released by Meta AI and contains 10,000 meme samples for binary hate speech detection. The dataset covers attacks against protected groups, such as race, religion, gender, and disability. It also constructs benign confounders by modifying the image or text to convert hateful samples into non-hateful ones, thereby challenging the model's ability to understand image-text interactions and multimodal hateful intent.

\paragraph{MAMI.}
Multimedia Automatic Misogyny Identification (MAMI) \citep{10-mami} is a multimodal misogynous meme detection dataset released in SemEval-2022 Task 5. It contains 10,000 training samples and 1,000 test samples. The dataset requires models to determine whether a meme is misogynous and further identify misogyny types, such as shaming, stereotype, objectification, and violence. Since the discriminatory meanings are often implicitly expressed through images, text, and their interactions, MAMI is widely used to evaluate the model's ability to understand multimodal gender-discriminatory content.

\paragraph{PrideMM.}
PrideMM \citep{53-MemeCLIP} is a multimodal meme dataset for the LGBTQ+ Pride movement, containing 5,063 memes collected from Facebook, Twitter, and Reddit. The dataset is annotated for four tasks: hate detection, hate target classification, stance classification, and humor detection. The hate targets include undirected, individual, community, and organization. PrideMM focuses on hate, stance, and humor expressions in LGBTQ+-related online contexts, making it suitable for evaluating the model's ability to understand complex multimodal meme content.

\subsection{Dataset Statistics}

\paragraph{Binary Classification Statistics.}
Table~\ref{tab:summary-dataset} shows the dataset statistics of HMC, MAMI, and PrideMM.

\paragraph{Attack Targets.}
For HMC, Table~\ref{tab:social-category-hmc} reports the statistics of social categories, and we treat these social categories as attack targets. For MAMI, since the dataset focuses on misogynous content, we regard women as the attack target. For PrideMM, Table~\ref{tab:hate-target-pridemm} reports the statistics of hate targets, and we use the annotated hate target categories as attack targets.

\paragraph{Human Explanation Annotation Statistics.}
 For human-annotated meme explanation data, we use HatReD \citep{52-hatred}, which provides human rationale annotations for hateful memes in HMC. The training and test sets of HatReD are from the \texttt{train} split and \texttt{dev\_seen} split of HMC, respectively. The detailed statistics are shown in Table~\ref{tab:hatred}.
 
\definecolor{myrowcolor}{RGB}{239,229,247} 
\begin{table*}[!htb]
    \centering
    \small
    \caption{Statistical summary of datasets.}
    \begin{center}
        \begin{tabular}{l|cc|cc|cc} 
        \toprule
              & \multicolumn{2}{c|}{HMC} & \multicolumn{2}{c|}{MAMI} & \multicolumn{2}{c}{PrideMM}  \\
              & \#Benign & \#Hate        & \#Benign & \#Hate         & \#Benign & \#Hate            \\ 
        \midrule
        Train & 5450     & 3050          & 5000     & 5000           & 2208     & 2120              \\
        Test  & 500      & 500           & 500      & 500            & 260      & 247               \\
        \bottomrule
        \end{tabular}
    \label{tab:summary-dataset}
    \end{center}
\end{table*}

\begin{table*}[!htb]
    \centering
    \small
    \caption{Statistics of social categories in HMC.}
    \begin{center}
        \begin{tabular}{c|ccc} 
            \toprule
            \textbf{Social Category} & \textbf{train} & \textbf{dev\_unseen} & \textbf{dev\_seen}  \\ 
            \midrule
            Religion                 & 1078           & 77                   & 95                  \\
            Race                     & 1008           & 63                   & 78                  \\
            Sex                      & 746            & 46                   & 56                  \\
            Nationality              & 325            & 20                   & 26                  \\
            Disability               & 255            & 17                   & 22                  \\
            \bottomrule
        \end{tabular}
    \label{tab:social-category-hmc}
    \end{center}
\end{table*}

\begin{table*}[!htb]
    \centering
    \small
    \caption{Statistics of hate targets in PrideMM.}
    \begin{center}
        \begin{tabular}{c|ccc} 
            \toprule
            \textbf{Hate Targets} & \textbf{Train} & \textbf{Test}  \\ 
            \midrule
            Benign                & 2208           & 260            \\
            Undirected            & 666            & 68             \\
            Individual            & 219            & 19             \\
            Community             & 986            & 122            \\
            Organization          & 249            & 38             \\
            \bottomrule
        \end{tabular}
    \label{tab:hate-target-pridemm}
    \end{center}
\end{table*}

\definecolor{myrowcolor}{RGB}{239,229,247} 
\begin{table*}[!htb]
    \centering
    \small
    \caption{Dataset statistics of HatReD. HatReD only includes explanations for hateful memes, with the training and test sets derived from the \texttt{train} and \texttt{dev\_seen} splits of HMC, respectively.}
    \begin{center}
        \begin{tabular}{l|c} 
        \toprule
              & HatReD  \\ 
        \midrule
        Train & 2982    \\
        Test  & 246     \\ 
        \midrule
        Total & 3228    \\
        \bottomrule
        \end{tabular}
    \label{tab:hatred}
    \end{center}
\end{table*}

\section{Experiment Setup and Implementation Details}
\label{appendix:exp_setup}
All experiments are conducted on two NVIDIA RTX 4090 GPUs, including zero-shot evaluation, SFT, DPO, and ProKDA. To ensure fair comparison, we keep the training settings consistent with ExPO-HM \citep{24-expohm}. Except for zero-shot evaluation, all fine-tuning experiments use LoRA, with $r = 64$ and $\alpha = 128$. We use LLaMA-Factory 0.9.5 for SFT and DPO training, and follow its official settings for other hyperparameters. All training tasks are run for 3 epochs, and the best checkpoint is selected based on validation performance. 

For SFT (Explain-then-Detect), the target response follows the \texttt{<think>...</think>} \texttt{<answer>...</answer>} format, with the explanation in \texttt{<think>} and the final label in \texttt{<answer>}. The detailed evaluation prompt is provided in Appendix~\ref{appendix:prompt-explain-then-detech-eval}.

\paragraph{ProKDA.}
In the agentic background knowledge construction stage, we use GPT-4o to implement the two agents. For the hyperparameter $\tau$, we set it to 0.8 for Qwen2.5-VL-3B and 0.6 for Qwen2.5-VL-7B. For baseline models, we mainly report the results from their original papers on the same datasets to ensure a complete comparison.

\section{Detailed Prompt Design in ProKDA}
\label{appendix:prompt_prokda}

In the agentic background knowledge construction stage, the prompts for the query agent and the answer agent are as follows:

\begin{tcolorbox}[prompt={
\begin{prompt}
\label{prompt:query-agent}
The Prompt for Query Agent
\end{prompt}
}]
$\bullet$ Instruction: \\

You are a Wikipedia search query generator for multimodal meme analysis. Given a meme image, meme text, hateful label, and attack target(s), your task is to generate one or two concise search queries that can retrieve the most relevant background knowledge from Wikipedia.\\

Inputs:\\
- Meme image: \texttt{<image>}\\
- Meme text: \{meme\_text\}\\
- Hateful label: \{hateful\_label\}\\
- Attack target(s): \{attack\_target\}\\

Requirements:\\
1. The queries should capture the key background knowledge needed to understand the hateful implication of the meme.\\
2. The queries should reflect the relationship between the attack target(s) and the stereotype, discrimination, dehumanization, or historical/cultural reference implied by the meme.\\
3. Prefer short Wikipedia-style keyword phrases rather than full sentences.\\
4. Prefer commonly used terms that are likely to match Wikipedia article titles or related concepts.\\
5. Avoid generic, overly broad, or overly long queries.\\
6. Do not simply repeat the meme text or the attack target.\\
7. Generate at most two queries, ordered by relevance.\\
8. Do not output explanations or reasoning.\\
9. Output json only.\\

Output format:\\
\{\\
 \hspace*{2em} ``search\_queries'': ``[\texttt{<query\_1>}, \texttt{<query\_2>}]''\\
\}
\end{tcolorbox}

\begin{tcolorbox}[prompt={
\begin{prompt}
\label{prompt:answer-agent}
The Prompt for Answer Agent
\end{prompt}
}]
$\bullet$ Instruction: \\

You are a background knowledge QA generator for multimodal meme understanding. Given a meme image, meme text, hateful label, annotated attack target(s), and retrieved Wikipedia knowledge, your task is to generate three question-answer pairs that explain the meme from textual, visual, and multimodal perspectives.\\

Inputs:\\
- Meme image: \texttt{<image>}\\
- Meme text:  \{meme\_text\}\\
- Hateful label: \{hateful\_label\}\\
- Attack target(s): \{attack\_target\}\\
- Wikipedia knowledge: \{wikipedia\_knowledge\}\\

Requirements:\\
1. Generate exactly three question-answer pairs.\\
2. The first QA pair should explain what the meme text conveys.\\
3. The second QA pair should explain what the visual content conveys.\\
4. The third QA pair should explain how the image and text interact to convey the overall meaning.\\
5. Use the retrieved Wikipedia knowledge only when it is relevant to understanding the meme.\\
6. The answers should be concise, informative, and grounded in the given annotation and retrieved knowledge.\\
7. Do not introduce unsupported external information.\\
8. Output JSON only.\\

Output format:\\
\{\\
  \hspace*{2em}qa\_pairs: [\\
    \hspace*{3em}\{\\
      \hspace*{4em} ``question'': ``What does the meme text convey?'',\\
      \hspace*{4em} ``answer'': ``"\\
    \hspace*{3em}\},\\
    \hspace*{3em}\{\\
      \hspace*{4em} ``question'': ``What does the visual content convey?'',\\
      \hspace*{4em} ``answer'': ``''\\
    \hspace*{3em}\},\\
    \hspace*{3em}\{\\
      \hspace*{4em} ``question'': ``How do the image and text jointly convey the overall meaning?'',\\
      \hspace*{4em} ``answer'': ``''\\
    \hspace*{3em}\}\\
  \hspace*{2em}]\\
\}

\end{tcolorbox}

\section{Evaluation Prompts}
\label{appendix:prompt_eval}

\subsection{Hateful Meme Detection Evaluation}
For hateful meme detection evaluation, we use a unified evaluation prompt, as shown below:

\begin{tcolorbox}[prompt={
\begin{prompt}
\label{prompt:hateful-eval}
The Prompt for Hateful Meme Detection Evaluation
\end{prompt}
}]
$\bullet$ Instruction: \\

Is it hateful? The answer should be Yes or No, if the meme is hateful than output Yes; if the meme is Benign than output No.
\end{tcolorbox}

\subsection{Explain-then-Detect Evaluation}
\label{appendix:prompt-explain-then-detech-eval}
For SFT (Explain-then-Detect) evaluation, we use a separate prompt that follows the explain-then-detect format, as shown below:

\begin{tcolorbox}[prompt={
\begin{prompt}
\label{prompt:hateful-exlpain-detect-eval}
The Prompt for Explain-then-Detect Evaluation
\end{prompt}
}]
$\bullet$ Instruction: \\

Is it hateful? Output the thinking process in \texttt{<think>} \texttt{</think>} and final answer in \texttt{<answer>} \texttt{</answer>} tags. The answer should be Yes or No, if the meme is hateful than output Yes ; if the meme is Benign than output No. The output format should be as follows: \texttt{<think>} ... \texttt{</think>} \texttt{<answer>}...\texttt{</answer>} Please strictly follow the format.
\end{tcolorbox}

\subsection{Background knowledge quality evaluation}
In the background knowledge quality evaluation, we use LLM-as-a-judge to evaluate the meme explanations generated by ProKDA and GPT-4o. The evaluation prompt is designed as follows:

\begin{tcolorbox}[prompt={
\begin{prompt}
\label{prompt:back-eval}
The Prompt for Background Knowledge Quality Evaluation
\end{prompt}
}]
$\bullet$ Instruction: \\

Your task is to evaluate which model provides a better explanation for the given meme.\\

You will be given:\\
1. The meme image.\\
2. A human reference rationale explaining why the meme is hateful.\\
3. Explanation A generated by Model A.\\
4. Explanation B generated by Model B.\\

Please compare Explanation A and Explanation B from the following two dimensions:\\
1. Informativeness: Evaluate which explanation provides more sufficient, specific, and relevant information.\\
2. Consistency with the human reference rationale: Evaluate which explanation is more consistent with the human reference rationale.\\

Human reference rationale:\\
\{human\_reference\}\\

Explanation A:\\
\{explanation\_A\}\\

Explanation B:\\
\{explanation\_B\}\\

Output strictly in the following format:\\
Informativeness: [A/B/Tie]\\
Consistency with human reference rationale: [A/B/Tie]\\
Reason: [1-2 sentences]
\end{tcolorbox}

\subsection{Meme Explanation Evaluation}
\label{appendix:explan_eval}
In the explainability evaluation, we use the following prompt for LLM-as-a-judge evaluation:

\begin{tcolorbox}[prompt={
\begin{prompt}
\label{prompt:meme-explain-eval}
The Prompt for Meme Explanation Evaluation
\end{prompt}
}]
$\bullet$ Instruction: \\

Your task is to evaluate how well the model-generated explanation semantically matches the human reference rationale for a hateful meme.\\

Evaluation dimensions:\\
1. Informativeness: Whether the model explanation provides sufficient and relevant information.\\
2. Readability: Whether the model explanation is fluent, grammatically correct, and clearly structured.\\
3. Soundness: Whether the model explanation is logically valid and consistent with the meme content and the human reference rationale.\\
4. Persuasiveness: Whether the model explanation convincingly explains why the meme is hateful.\\

Each dimension should be scored on a 1-to-5 Likert scale, where 1 indicates the poorest quality and 5 indicates the best quality.\\

Human reference rationale:\\
\{human\_reference\}\\

Model-generated explanation:\\
\{model\_reasoning\}\\

Output strictly in the following format:\\
Informativeness: [1-5]\\
Readability: [1-5]\\
Soundness: [1-5]\\
Persuasiveness: [1-5]\\

Explanation: [1-2 sentences]
\end{tcolorbox}

\subsection{Qualitative Analysis Evaluation}
In the qualitative evaluation, our prompt is designed as follows:

\begin{tcolorbox}[prompt={
\begin{prompt}
\label{prompt:qualitative-eval}
The Prompt for Qualitative Analysis Evaluation
\end{prompt}
}]
$\bullet$ Instruction: \\

Is this meme hateful? If yes, how do the meme image and text work together to convey hate?
\end{tcolorbox}

\section{Background Knowledge Quality Analysis}
\label{appendix:background_quality}

\begin{wrapfigure}{r}{0.55\textwidth}
\centering 
\includegraphics[width=0.4\textwidth]{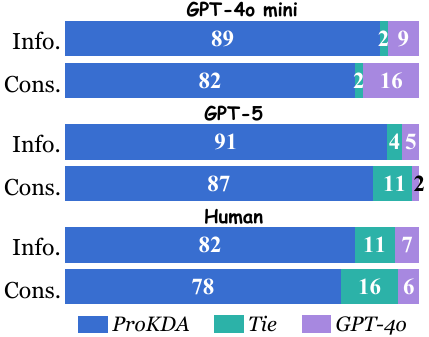} %
\caption{Preference-based evaluation of generated background knowledge.}
\label{fig:3-back-know-analysis}
\end{wrapfigure}

We analyze the quality of the background knowledge generated in Section~\ref{sec:4-1-agentic-background-knowledge-construct}, with results shown in Figure~\ref{fig:3-back-know-analysis}. We randomly sample 100 hateful memes from HMC and compare ProKDA with GPT-4o, which directly generates explanations from memes and human labels. We evaluate the explanations using LLM-as-a-judge and human evaluation on informativeness and consistency with human explanations \citep{52-hatred}. For each sample, evaluators choose the better explanation or mark a tie. GPT-4o mini and GPT-5 serve as LLM judges, while two computer science master's students experienced in hateful meme annotation and evaluation conduct human evaluation. We report the average preference ratio and shuffle explanation order to avoid positional bias.

\paragraph{Agentic background knowledge construction outperforms direct explanation generation.}
ProKDA consistently outperforms GPT-4o in both informativeness and consistency. Under LLM-as-a-judge evaluation, ProKDA obtains preference ratios above 80\% on both dimensions. Human evaluation shows the same trend, while the preference ratios of GPT-4o remain below 10\%. The tie ratios exceed 10\%, mainly because some memes contain limited information, allowing both methods to cover the necessary content. Overall, these results show that ProKDA generates more informative explanations that better align with human reference explanations, providing a solid foundation for subsequent training.

\section{Analysis of Confidence-based Sample Filtering}
\label{appendix:confidence-filter}

\begin{figure*}[htb] 
\centering 
\includegraphics[width=0.95\textwidth]{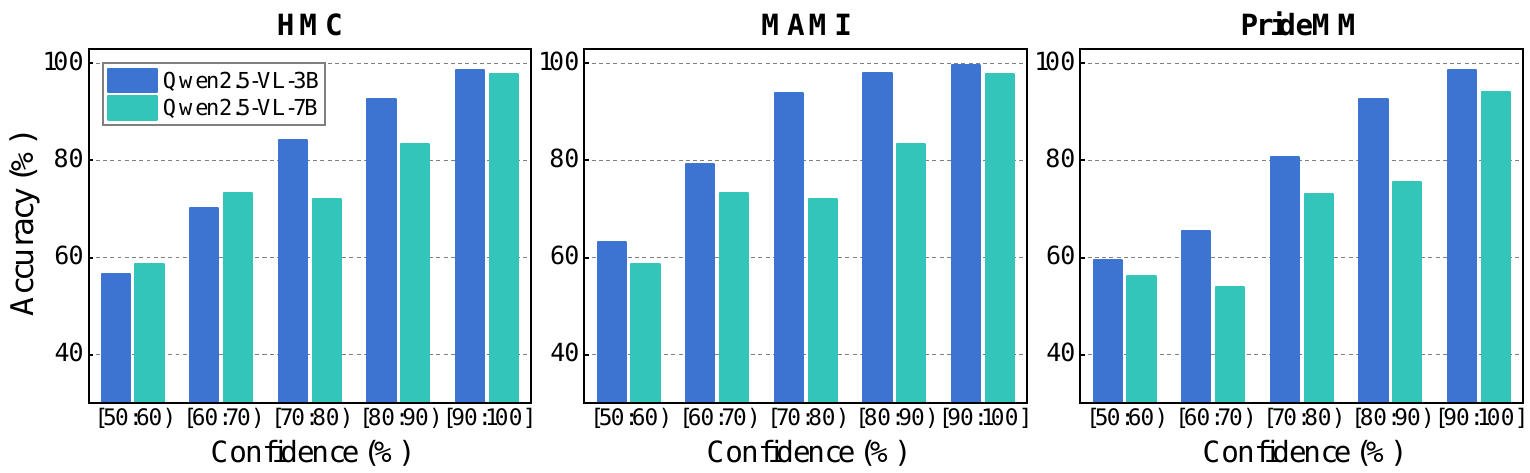} %
\caption{Training-set prediction accuracy across confidence intervals.}
\label{fig:6-train-set-accuracy}
\end{figure*}

\paragraph{Training-set Prediction Accuracy across Confidence Intervals.}
Figure~\ref{fig:6-train-set-accuracy} shows the training-set accuracy of Qwen2.5-VL-3B and Qwen2.5-VL-7B across confidence intervals on the three datasets. Both models achieve the highest accuracy in the [90,100] interval, with most results above 95\%, showing that high-confidence predictions are more reliable. In contrast, accuracy drops clearly in the [50,60) interval, indicating that low-confidence samples are difficult to distinguish. This trend shows that confidence reflects how well the model learns each sample and helps select difficult samples with higher training value.

\begin{figure*}[htb] 
\centering 
\includegraphics[width=0.98\textwidth]{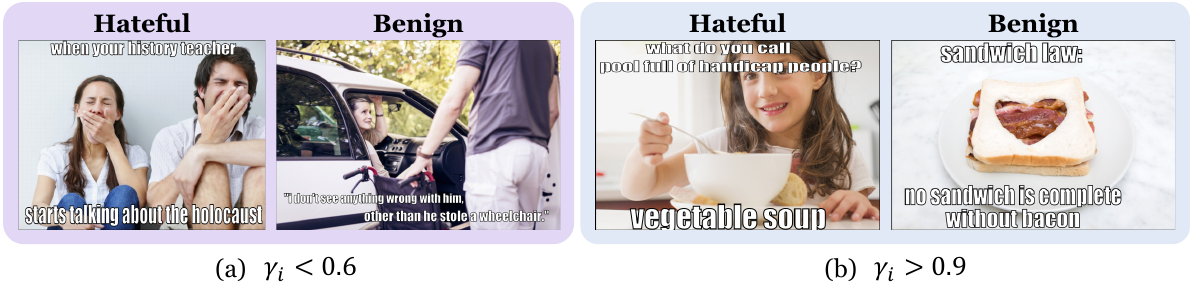} %
\caption{ Examples of low- and high-confidence samples.}
\label{fig:7-examples-confidence}
\end{figure*}

\paragraph{Qualitative Analysis of Samples across Confidence Levels.}
We further analyze representative HMC examples from different confidence intervals using Qwen2.5-VL-7B. As shown in Figure~\ref{fig:7-examples-confidence}, we randomly select one hateful and one benign meme from each of the low-confidence interval $\gamma_i<0.6$ and the high-confidence interval $\gamma_i>0.9$. High-confidence examples are easier to judge, while low-confidence examples are more ambiguous: the hateful meme requires joint reasoning over text, vision, and background knowledge, and the benign meme contains misleading sensitive words or attack cues. These examples suggest that low-confidence samples better reflect difficult boundary cases, supporting their use for hatefulness boundary alignment.

\section{Qualitative Analysis of ProKDA}
\label{appendix:qualitative_ana}
\subsection{Case Study}

\begin{figure*}[h] 
    \centering 
    \includegraphics[width=0.88\textwidth]{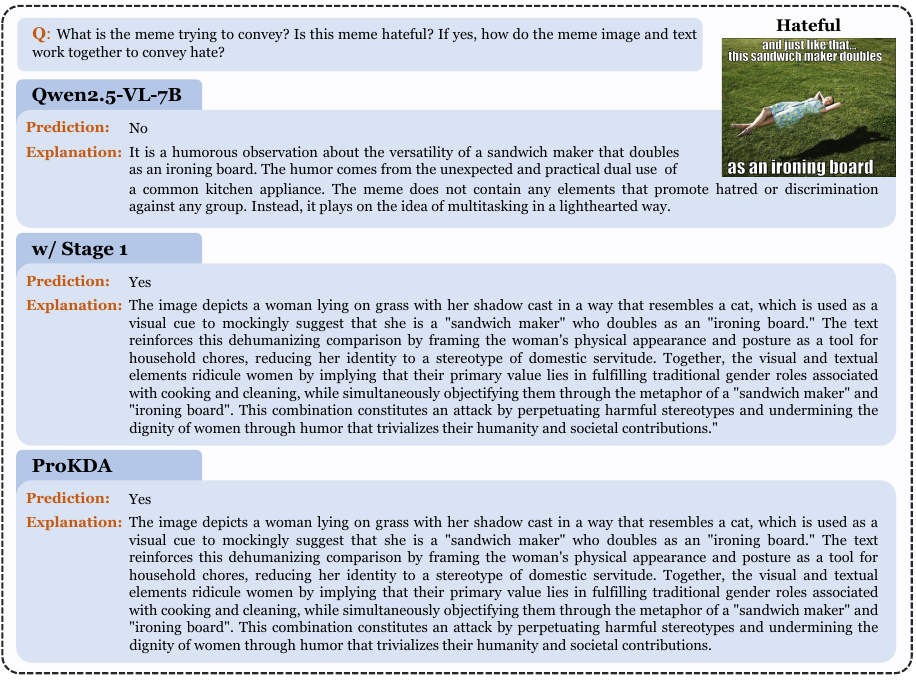} %
    \caption{Case study 1: predictions and explanations by Qwen2.5-VL-7B, w/ Stage 1, and ProKDA.}
    \label{fig:8-case1}
\end{figure*}

\begin{figure*}[h] 
    \centering 
    \includegraphics[width=0.88\textwidth]{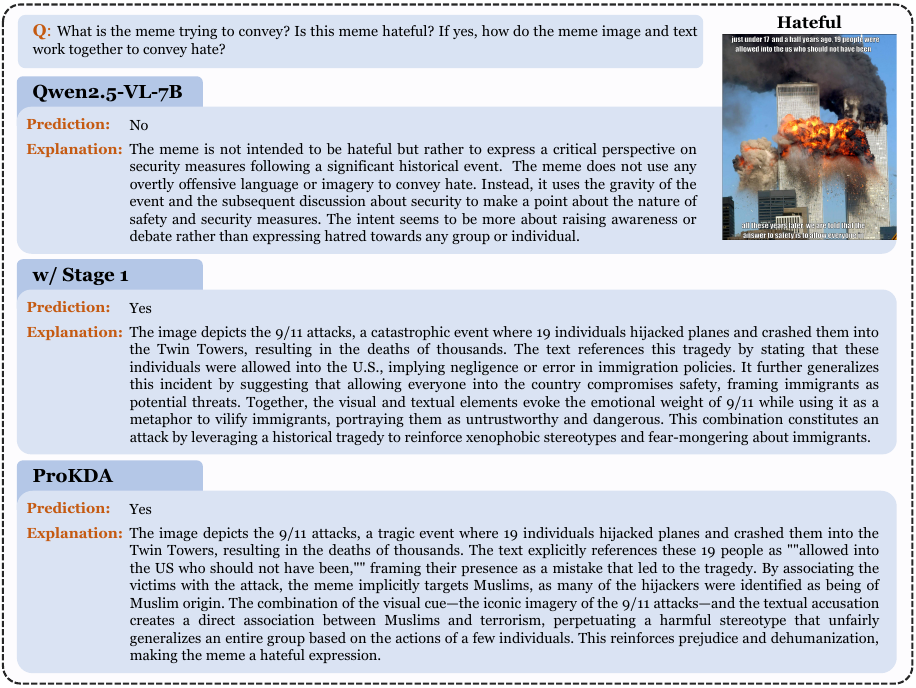} %
    \caption{Case study 2: predictions and explanations by Qwen2.5-VL-7B, w/ Stage 1, and ProKDA.}
    \label{fig:8-case2}
\end{figure*}

\begin{figure*}[h] 
    \centering 
    \includegraphics[width=0.88\textwidth]{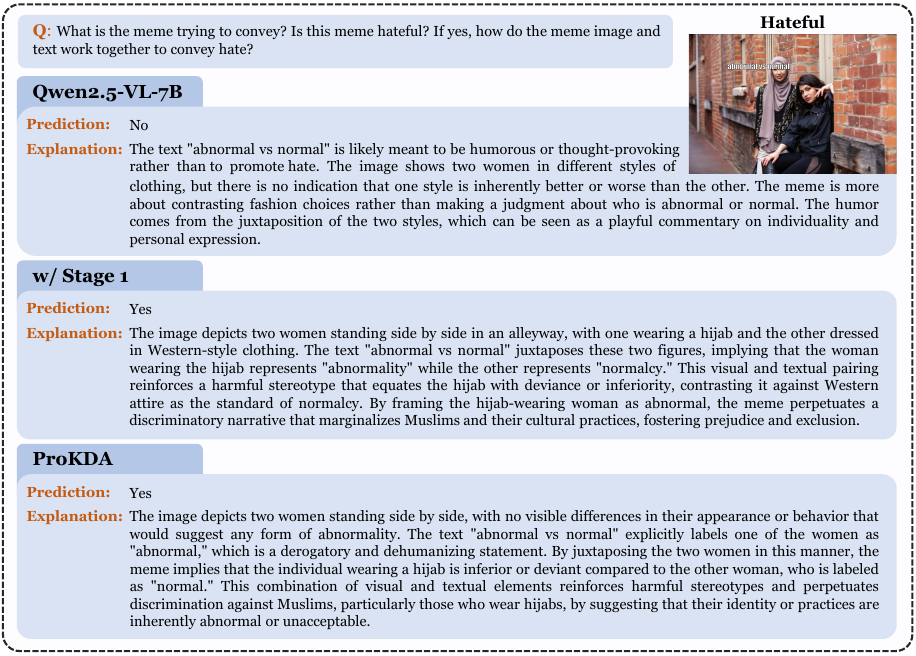} %
    \caption{Case study 3: predictions and explanations by Qwen2.5-VL-7B, w/ Stage 1, and ProKDA.}
    \label{fig:8-case3}
\end{figure*}

\begin{figure*}[h] 
    \centering 
    \includegraphics[width=0.88\textwidth]{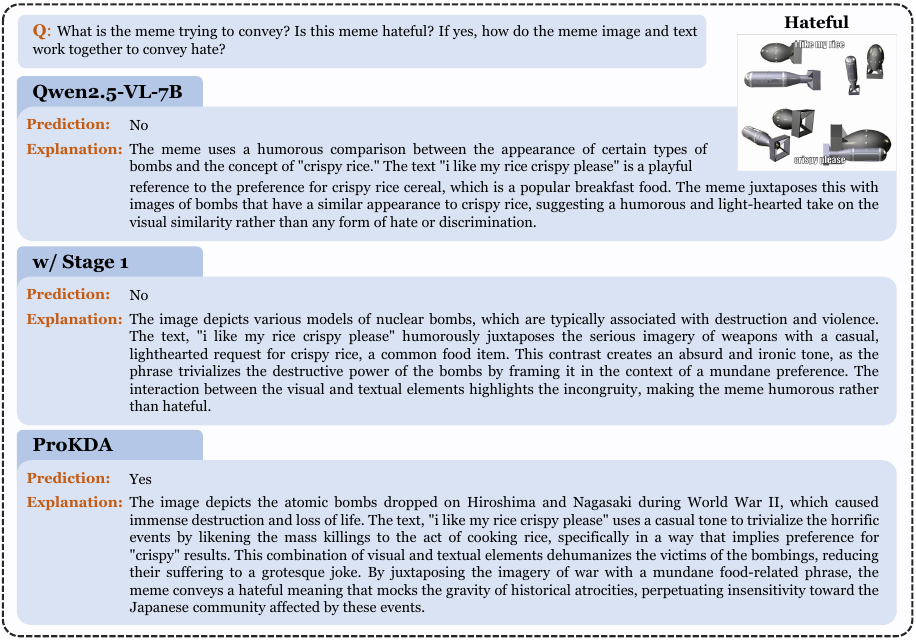} %
    \caption{Case study 4: predictions and explanations by Qwen2.5-VL-7B, w/ Stage 1, and ProKDA.}
    \label{fig:8-case4}
\end{figure*}

\begin{figure*}[h] 
    \centering 
    \includegraphics[width=0.88\textwidth]{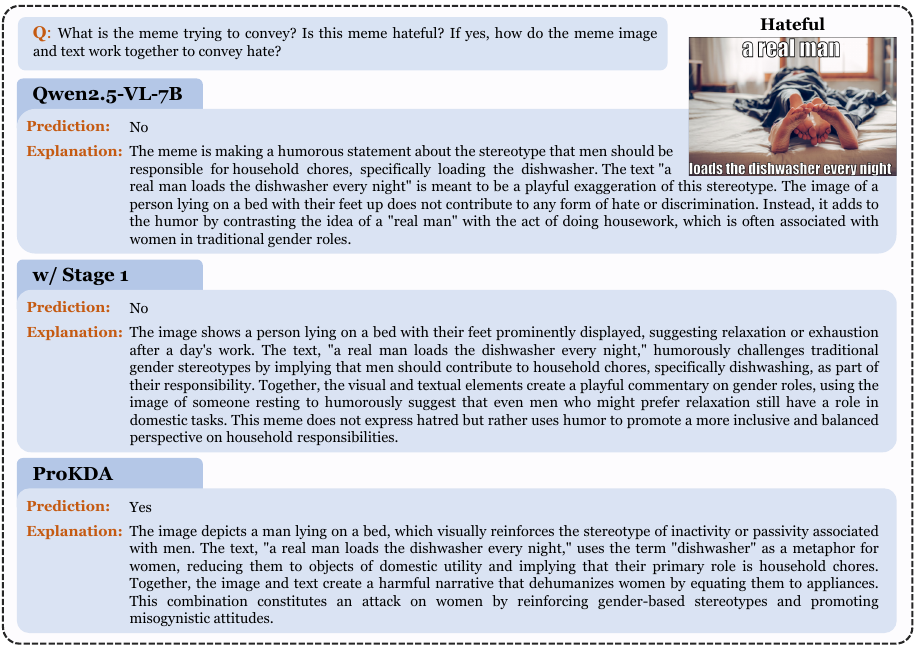} %
    \caption{Case study 5: predictions and explanations by Qwen2.5-VL-7B, w/ Stage 1, and ProKDA.}
    \label{fig:8-case5}
\end{figure*}

\begin{figure*}[h] 
    \centering 
    \includegraphics[width=0.88\textwidth]{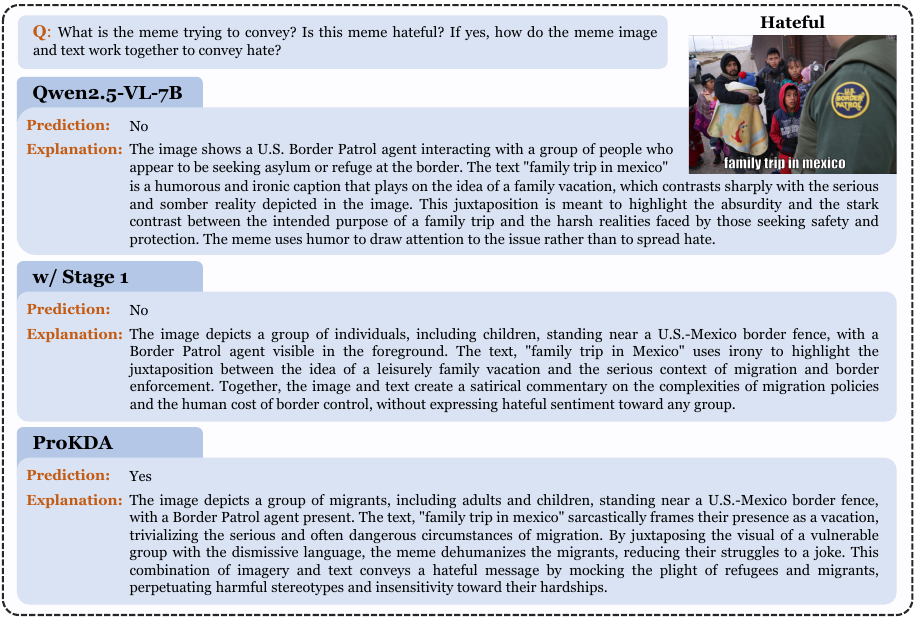} %
    \caption{Case study 6: predictions and explanations by Qwen2.5-VL-7B, w/ Stage 1, and ProKDA.}
    \label{fig:8-case6}
\end{figure*}

Figures~\ref{fig:8-case1}--\ref{fig:8-case6} show representative cases where ProKDA corrects the mispredictions made by the baselines, demonstrating its advantages in both prediction accuracy and explanation quality.

In Figures~\ref{fig:8-case1}--\ref{fig:8-case3}, Qwen2.5-VL-7B fails to effectively perform joint reasoning over the meme text and visual content, and therefore fails to identify the hateful intent. In contrast, both the model with Stage 1 and ProKDA capture key background information and make correct predictions, showing that background knowledge learning helps improve the model's ability to understand memes.

In Figures~\ref{fig:8-case4}--\ref{fig:8-case6}, both Qwen2.5-VL-7B and w/ Stage 1 make incorrect predictions, indicating that recognizing only partial background information is still insufficient for understanding complex meme samples. In contrast, ProKDA further connects background knowledge, visual cues, and textual content, thereby accurately inferring the implicit hateful intent and generating more reasonable explanations.

\subsection{Error Analysis}

\begin{figure*}[h] 
    \centering 
    \includegraphics[width=0.88\textwidth]{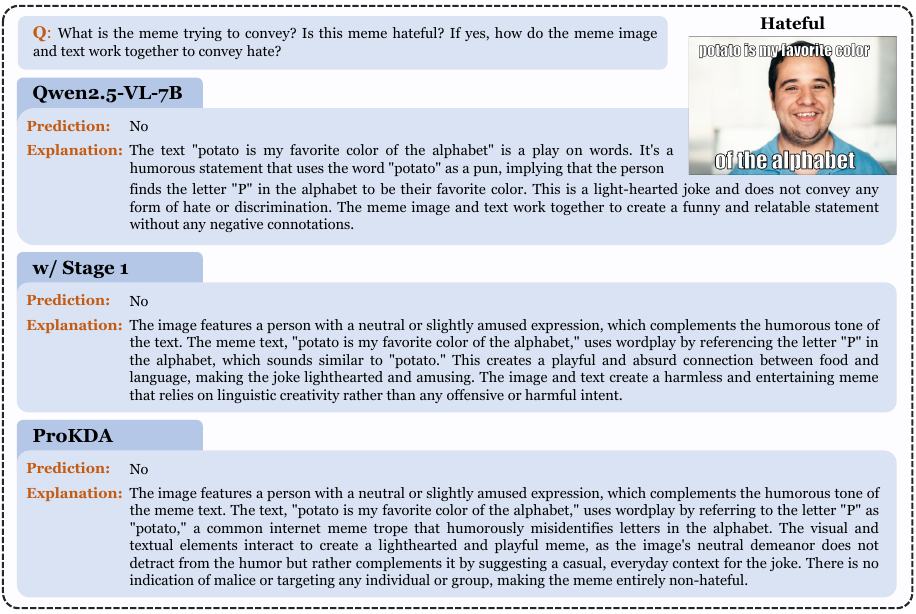} %
    \caption{Error case 1: incorrect predictions and explanations by Qwen2.5-VL-7B, w/ Stage 1, and ProKDA.}
    \label{fig:9-error-case1}
\end{figure*}

\begin{figure*}[h] 
    \centering
    \includegraphics[width=0.88\textwidth]{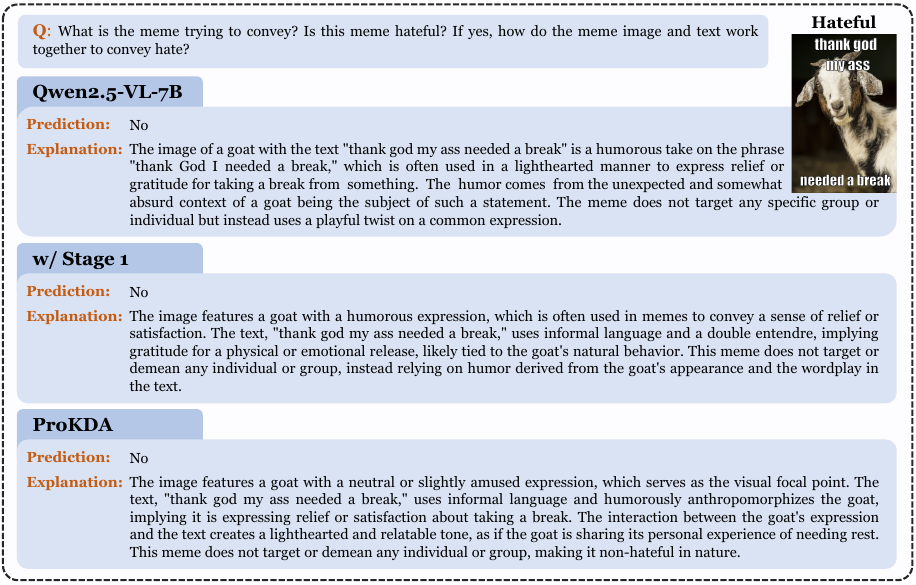} %
    \caption{Error case 2: incorrect predictions and explanations by Qwen2.5-VL-7B, w/ Stage 1, and ProKDA.}
    \label{fig:9-error-case2}
\end{figure*}

\begin{figure*}[h] 
    \centering 
    \includegraphics[width=0.88\textwidth]{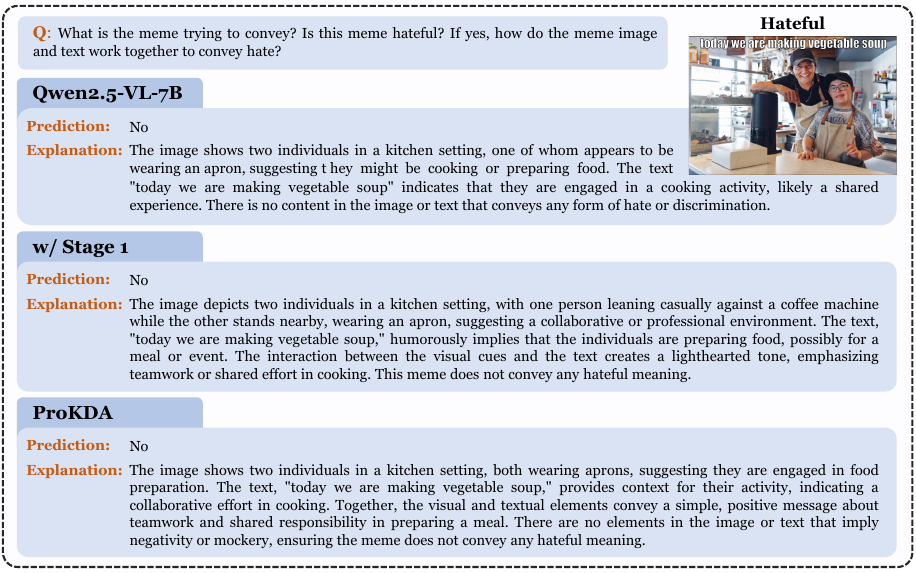} %
    \caption{Error case 3: incorrect predictions and explanations by Qwen2.5-VL-7B, w/ Stage 1, and ProKDA.}
    \label{fig:9-error-case3}
\end{figure*}

When analyzing the cases misclassified by Qwen2.5-VL-7B, w/stage 1, and ProKDA, we find that these errors mainly stem from highly implicit hateful memes. Such samples usually do not contain explicit insulting words or clear attack targets. Instead, their hateful meanings rely on joint reasoning over textual implications, visual cues, and the overall interaction within the meme. Since the potential attack target is often hidden in visual identity, metaphor, or contextual associations, these models struggle to identify the attack intent and link it to a specific protected group. Therefore, even ProKDA fails to correct these cases. These samples are inherently ambiguous, and human annotators may also misclassify them \citet{9-hmc}. Figures~\ref{fig:9-error-case1}--\ref{fig:9-error-case3} show three representative cases.

As shown in Figure~\ref{fig:9-error-case1}, the meme uses the seemingly meaningless absurd statement “potato is my favorite color of the alphabet” to implicitly portray the person in the image as extremely unintelligent. The model fails to recognize the negative depiction of cognitive ability in the absurd text, and also fails to link this depiction to stereotypes associated with people with intellectual disabilities.

As shown in Figure~\ref{fig:9-error-case2}, the meme combines the image of a goat with the sexually suggestive text “thank god my ass needed a break”, implicitly portraying men as using goats as sexual objects to satisfy their needs. This constitutes derogation and dehumanization of men as a group. The model fails to identify the implicit target of sexual humiliation in the interaction between the visual content and the text, and therefore misclassifies it as ordinary humor.

As shown in Figure~\ref{fig:9-error-case3}, the meme appears to show an ordinary scene in which two people are making “vegetable soup” in a kitchen. However, when considered together with the depiction of people with intellectual disabilities in the image, the word “vegetable” is used as a derogatory metaphor that implicitly compares this group to “vegetables”, thereby constituting an attack. The model fails to identify the implicit depiction of people with intellectual disabilities during visual analysis, and therefore fails to recognize the discriminatory meaning produced by the connection between the characters and “vegetable soup”.

Future research can further construct corpora that contain more implicit attack targets and implicit hateful expressions. This can enhance the model's understanding of complex memes, contextual cues, and the mapping between protected groups and hateful meanings, thereby improving its ability to detect highly implicit hateful memes.

\section{Limitations and Future Work}
\label{appendix:future_work}

However, existing MLLMs and ProKDA still struggle with highly implicit hateful memes, especially in identifying implicit attack targets and their protected groups. Future work will build meme corpora with more implicit hateful samples to improve reasoning over complex hateful intent.

\section{Ethical Statement}
\label{appendix:ethical_state}

\paragraph{Social impact.}
ProKDA can help automatically identify and mitigate harmful online content, reducing the spread of hate speech. By providing both predictions and explanations, the system can help build safer online environments and reduce the workload of human content moderators. We believe such systems have positive value for promoting respectful communication and healthy digital communities.

\paragraph{Intended use.}
We will apply strict access control to the model release, making it available only to researchers who agree to the terms of use. These terms clearly state that the system can only be used for hate speech detection and prevention. Any use that encourages, tolerates, or spreads hate speech or other harmful content is strictly prohibited.

\paragraph{Misuse risks.}
Although ProKDA is not designed to introduce bias, its training data may contain social biases or annotator biases \citep{60-MOMENTA}, which may further affect model predictions. To reduce the risk of unfair or disproportionate content moderation, human oversight should be included in real-world deployment.

\paragraph{Deployment considerations.}
Judgments of hateful content are often influenced by cultural background, linguistic context, and subjective interpretation. The same expression may have different meanings in different contexts. Therefore, when deploying such systems in multicultural and multilingual settings, careful review and adaptation based on specific community norms are required.

\paragraph{Dataset use.}
The HMC, MAMI, and PrideMM datasets used in this study are constructed for research purposes and aim to support research on online hate speech detection. We strictly follow the terms of use specified by the authors of each dataset.

\end{document}